\documentclass[10pt,twocolumn,letterpaper]{article}

\usepackage[wacvfinal,algorithms]{wacv}      % To produce the REVIEW version for the algorithms track
\usepackage{graphicx}
\usepackage{amsmath}
\usepackage{amssymb}
\usepackage{booktabs}

\usepackage[pagebackref,breaklinks,colorlinks]{hyperref}

\usepackage[capitalize]{cleveref}
\crefname{section}{Sec.}{Secs.}
\Crefname{section}{Section}{Sections}
\Crefname{table}{Table}{Tables}
\crefname{table}{Tab.}{Tabs.}

\usepackage{multirow}
\usepackage{colortbl}

\definecolor{cvprblue}{rgb}{0.21,0.49,0.74}
\definecolor{cvprgreen}{rgb}{0.10, 0.52, 0.27}
\definecolor{cvprgrey}{rgb}{0.5, 0.52, 0.5}
\definecolor{darkblue}{RGB}{0,0,180} % dark blue color
\definecolor{darkred}{RGB}{180,0,0} % dark red color
\definecolor{darkpurple}{RGB}{120,0,180} % dark purple color
\definecolor{darkgreen}{RGB}{0,120,0} % dark green color
\definecolor{darkbrown}{RGB}{100,40,0} % dark brown color
\definecolor{gray}{rgb}{0.5,0.5,0.5}
\definecolor{gray94}{gray}{.92}
\definecolor{gray90}{gray}{.90}
\definecolor{gray85}{gray}{.85}

\usepackage{pifont}% 
\usepackage{xspace}
\usepackage{placeins}
\newcommand{\cmarkg}{\textcolor{gray}{\ding{51}}\xspace}%
\newcommand{\xmarkg}{\textcolor{gray}{\ding{55}}\xspace}%
\def\wacvPaperID{*****} % *** Enter the WACV Paper ID here
\def\confName{WACV}
\def\confYear{2025}

\begin{document}

% \title{\LaTeX\ Author Guidelines for \confName~Proceedings}
\title{Multiple Contexts and Frequencies Aggregation Network for\\ Deepfake Detection}

\author{Zifeng Li\\
Beihang University\\
{\tt\small lizifeng@buaa.edu.cn}
\and
Wenzhong Tang\\
Beihang University\\
{\tt\small tangwenzhong@buaa.edu.cn}
\and
Shijun Gao\\
Beihang University\\
{\tt\small gaoshijun@buaa.edu.cn}
\and
Shuai Wang\\
Beihang University\\
{\tt\small wangshuai@buaa.edu.cn}
\and
Yanxiang Wang$^*$\\
Beihang University\\
{\tt\small wangyanyang@buaa.edu.cn}
\thanks{Yanxiang Wang is the corresponding author.}
}
\maketitle
%%%%%%%%% BODY TEXT
\begin{abstract}

Deepfake detection faces increasing challenges since the fast growth of generative models in developing massive and diverse Deepfake technologies.
Recent advances rely on introducing heuristic features from spatial or frequency domains rather than modeling general forgery features within backbones.
To address this issue, we turn to the backbone design with two intuitive priors from spatial and frequency detectors, \textit{i.e.,} learning robust spatial attributes and frequency distributions that are discriminative for real and fake samples.
To this end, we propose an efficient network for face forgery detection named MkfaNet, which consists of two core modules.
For spatial contexts, we design a Multi-Kernel Aggregator that adaptively selects organ features extracted by multiple convolutions for modeling subtle facial differences between real and fake faces. For the frequency components, we propose a Multi-Frequency Aggregator to process different bands of frequency components by adaptively reweighing high-frequency and low-frequency features.
Comprehensive experiments on seven popular deepfake detection benchmarks demonstrate that our proposed MkfaNet variants achieve superior performances in both within-domain and across-domain evaluations with impressive efficiency of parameter usage.

\end{abstract}
    
\begin{figure}[t!]
    \vspace{-1.0em}
    \centering
    \includegraphics[width=1.\linewidth]{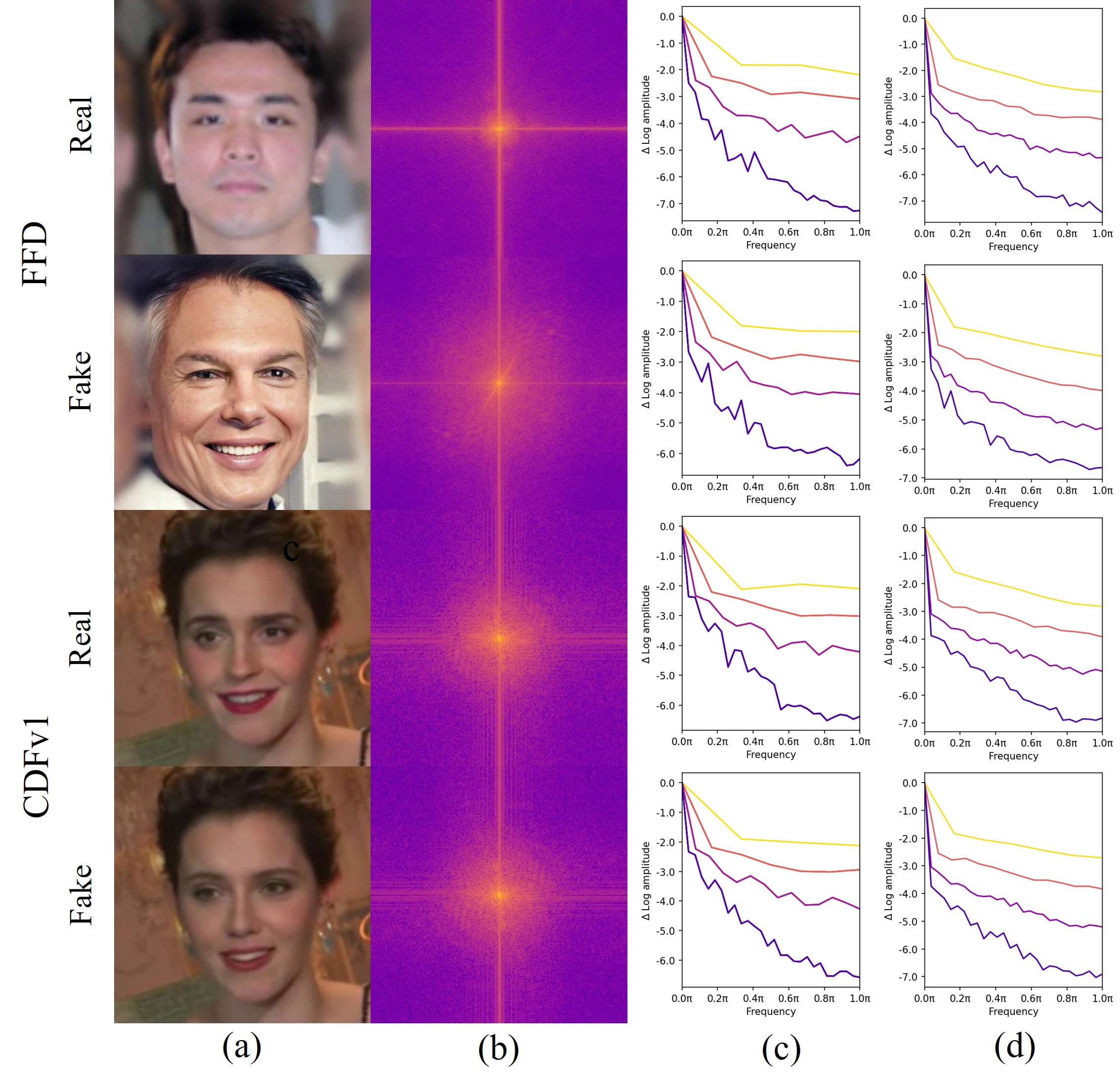}
    \vspace{-2.0em}
    \caption{
    % Face authenticity recognition.
    Illustration of frequency priors in deepfake detection.
    (a): Source image. (b): Data frequency domain analysis.  (c): Relative log amplitudes of Fourier transformed feature maps of ResNet50. (d): Relative log amplitudes of Fourier transformed feature maps of MkfaNet. (b) reveals the uniformity of the frequency distribution in real faces and the concentration of high-frequency anomalies in forged faces. (c) shows that ResNet50 has a relatively low logarithmic amplitude in the high-frequency region, indicating its insufficiency in capturing high-frequency details. (d) demonstrates that MkfaNet has a higher amplitude in the high-frequency region with broader coverage, highlighting its advantages in handling high-frequency details and identifying forgery features.}
    \label{fig:motivation}
    \vspace{-1.0em}
\end{figure}

\section{Introduction}
\label{sec:intro}
% background and danger of deepfake
With the development of generative models, Deepfake technology has made significant progress. Deepfake encompasses video, audio, and text, utilizing advanced artificial intelligence techniques such as Variational Autoencoders (VAE)~\cite{kingma2019introduction}, Generative Adversarial Networks (GAN)~\cite{creswell2018generative}, and Diffusion Models (DM)~\cite{croitoru2023diffusion} to achieve unprecedented realism. Unfortunately, these fake visual data can be used for malicious purposes, such as invading personal privacy, spreading misinformation, and undermining people's trust in digital media~\cite{cahlan2020misinformation, aaa}.
Considering that facial deepfakes can potentially cause more significant social and ethical implications compared to synthetic media without facial content, 
we specifically concentrate on facial deepfake technology in this paper.

To address the potential risks posed by Deepfakes, numerous researchers are working to enhance Deepfake detection technology and strengthen existing detection systems~\cite{le2023quality,feng2023self,bai2023aunet, ge2018low, ge2017detecting}.
These methods employ various techniques and are generally classified into three types: naive detectors~\cite{afchar2018mesonet,rossler2019faceforensics++}, spatial detectors~\cite{cao2022end,wang2021representative}, and frequency detectors~\cite{luo2021generalizing,qian2020thinking}.
Meanwhile, researchers are striving to develop sufficiently robust detectors to withstand various forms of degradation, such as noise~\cite{jiang2020deeperforensics,haliassos2021lips,li2022selective}, compression~\cite{le2023quality,woo2022add}, and, most critically, to identify previously unseen Deepfakes~\cite{pu2021deepfake,shiohara2022detecting}.
Therefore, enhancing the generalization ability of Deepfake detection models becomes particularly important. Models with strong generalization capabilities can effectively identify and counter new Deepfake attacks that have not appeared in the training data, thereby ensuring the authenticity and security of information~\cite{le2024sok}.

Improving the model's ability to capture critical facial features is an effective means of enhancing its generalization capability. These key features include but are not limited to, subtle dynamics of facial expressions, natural gradients of skin tone, and natural eye blinking. By accurately capturing these difficult-to-simulate details, the model can more effectively distinguish between real content and Deepfake-generated content~\cite{haliassos2021lips}. 
In recent research, adopting multitask learning~\cite{chen2022self,li2020face,chen2020learning,zhao2021learning} and/or heuristic fake data generation strategies~\cite{li2020face,sheng2023structure} is the mainstream method to enhance the generalization capability of Deepfake detection methodd. These approaches aim to improve the model's adaptability and discrimination ability against novel forgery techniques by learning multiple related tasks simultaneously. Meanwhile, heuristic data generation methods create new and unseen fake samples to test and improve the robustness of detection algorithms.
However, commonly used architectures for these methods, such as XceptionNet~\cite{chollet2017xception} and EfficientNet~\cite{tan2019efficientnet}, primarily tend to learn global features while neglecting more local features~\cite{wang2023dynamic,zhao2021multi, zhao2021graph}. Consequently, most of these methods fail to effectively model local artifacts, which is crucial for detecting high-quality Deepfake content. 

We first focus on the differences between real and forged samples in the frequency domain, and our empirical analysis reveals significant disparities in their frequency distributions, as shown in Figure~\ref{fig:motivation}(b). Specifically, real samples exhibit a relatively uniform energy distribution in the spectrogram, indicating a balanced texture and edge information across various frequencies. In contrast, forged samples display abnormally concentrated energy peaks in the high-frequency region, highlighting the shortcomings of forgery techniques in handling high-frequency details, which result in unnatural textures and edges in the high-frequency area. To further illustrate this phenomenon, we use a pre-trained ResNet50 model to examine how it processes real and fake face images, with the results shown in Figure~\ref{fig:motivation}(c). Notably, ResNet50 exhibits a weaker response in the high-frequency region, indicating its insufficiency in capturing the high-frequency details of forged faces. Additionally, when processing forged face images, ResNet50's shallow feature maps exhibit higher low-frequency responses, whereas these low-frequency responses are weaker and more uniformly distributed when processing real faces. This indicates that ResNet50 has a stronger reaction to simple features in forged images but lacks sensitivity to high-frequency details. This layered difference in frequency response reveals the underlying mechanisms by which deep networks distinguish between real and fake faces. It provides important insights and motivation for designing models that can more accurately differentiate between genuine and forged faces.

Additionally, we have observed that recent advances rely on introducing heuristic features from either the spatial or frequency domain, rather than establishing a general forgery feature detection model within the backbone network. While this approach improves detection performance to some extent, it still has limitations, especially in addressing the continuously evolving forgery techniques. Therefore, we propose MkfaNet, which integrates more powerful feature capture and analysis capabilities into the backbone network by combining the Multi-Kernel Aggregator (MKA) and Multi-Frequency Aggregator (MFA), significantly enhancing the accuracy and robustness of forgery detection.
In specific, the Multi-Kernel Aggregator (MKA) module combines depth-wise separable convolutions with different dilation rates to effectively expand the model's receptive field, enhancing its ability to capture features at various scales from the input data. It then adaptively selects features extracted through multiple convolutions based on the spatial context to model the subtle facial differences between real and fake faces;
Multi-Frequency Aggregator (MFA) module optimizes the model's response to different frequency information by separately processing and fusing the DC (Direct Current) and HC (High Current) components of images. MkfaNet, as a stack of MKA and MFA modules, shows the enhanced ability to discern image details and structural information. In the context of real and fake face recognition, it can accurately distinguish the subtle texture and frequency distortions introduced by forgery techniques, thereby improving the accuracy of fake image detection. 

Comprehensive experiments on seven popular deepfake detection benchmarks~\cite{yan2023deepfakebench} demonstrate that our proposed MkfaNet variants achieve superior performances in both within-domain and across-domain evaluations with impressive efficiency of parameter usage.

\begin{figure*}[!t] 
\centering
\includegraphics[width=1.01\textwidth]{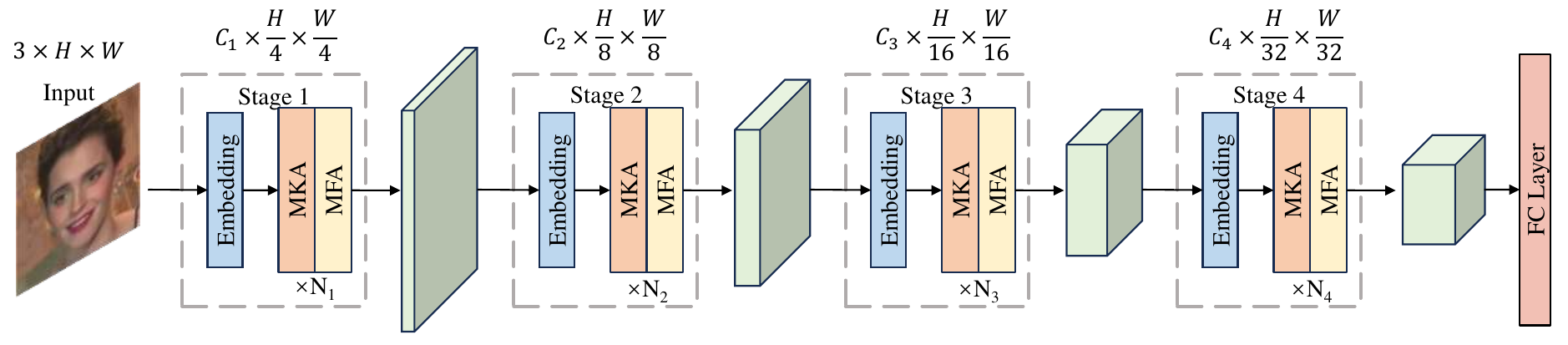} 
\caption{
MkfaNet architecture with four stages. MkfaNet uses a hierarchical architecture of 4 stages. Each stage $i$ consists of an embedding stem, $N_i$ Multi-Kernel Aggregator (MKA), and Multi-Frequency Aggregator (MFA) Blocks.} 
\label{fig:architecture} 
\vspace{-1.0em}
\end{figure*}
\section{Related Work}
\label{sec:related}

\paragraph{Deepfake Generation}
Deepfake technology primarily involves the artificial modification of facial images and has significantly evolved since its inception. Since 2017, machine learning-based facial manipulation techniques have made substantial advancements, particularly in the areas of facial replacement and facial expression reenactment, which have garnered widespread attention~\cite{yan2023deepfakebench}. Ian Goodfellow et al. introduced Generative Adversarial Networks (GANs)~\cite{goodfellow2020generative}, a technology that has significantly advanced the development of realistic image synthesis, including facial images~\cite{choi2018stargan,karras2017progressive}.
GANs consist of two parts: the generator and the discriminator. The generator is responsible for creating images, while the discriminator's task is to distinguish between these generated images and real data. Variational Autoencoders (VAEs)~\cite{kingma2013auto} compress data into a compact form and are used in Deepfake technology to alter facial features, such as expressions and styles.
Diffusion models (DMs)~\cite{zhao2023diffswap,gerogiannis2024animateme} create images by gradually adding noise and then progressively removing this noise during the generation process. In facial image generation, diffusion models can produce high-quality, high-resolution facial images by finely controlling the noise reduction process.
Facial Deepfakes can be broadly categorized into two types: face-swapping and face-reenactment. Face-swapping refers to replacing the facial features in one image with the facial features from another image~\cite{li2019faceshifter,nirkin2019fsgan,perov2020deepfacelab}.
Face-reenactment technology modifies the original face using image processing techniques to mimic the expressions of another face. Face2Face~\cite{thies2016face2face} generates different expressions by tracking facial key points, while NeuralTextures~\cite{thies2019deferred} achieves expression transfer using rendered images generated from 3D facial models. These technologies enable more diverse and precise simulation of facial expressions.

\vspace{-10pt}

\paragraph{Deepfake Detection}
In Deepfake detection research, there are two main types: image-level detectors and video-level detectors. Image-level detectors identify fake images by recognizing spatial artifacts in single frames.
A direct method for detecting spatial artifacts is using the Xception model~\cite{rossler2019faceforensics++}, a convolutional neural network (CNN) architecture, combined with the use of attention mechanisms like MAT~\cite{zhao2021multi}.
Face X-ray~\cite{li2020face} proposed a method that uses boundaries between forged faces and backgrounds to capture spatial inconsistency.
Recently, algorithms utilize methods including blending artifacts~\cite{bai2023aunet,dong2023implicit} or separating elements relevant to detection from irrelevant ones during their training.
In contrast to image-level detectors, video-level detectors take advantage of temporal information by using multiple frames to detect deepfake videos~\cite{choi2024exploiting}.
Recently, FTCN~\cite{zheng2021exploring} directly extracted temporal information using 3D CNNs with a spatial kernel size of 1. AltFreeze~\cite{wang2023altfreezing} showcases strong generalization capabilities through its unique approach of independently training spatial and temporal information.
Despite their great potential, the aforementioned models are less robust when considering high-quality deepfakes. Indeed, these SoA methods mainly employ traditional DNN backbones such as XceptionNet~\cite{chollet2017xception} and EfficientNet~\cite{tan2019efficientnet}. 

Therefore, these network models implicitly form global features through their successive convolutional layers. This can lead to some critical and useful features being unintentionally overlooked, thereby affecting the effective detection of high-quality Deepfake content. Thus, it is essential to develop effective strategies that focus on capturing key features to achieve high-quality and effective detection.

\section{Method}

% \subsection{Motivation}
% A key contributing factor to the generalization problem in deepfake detection is the issue of overfitting to forgery-specific artifacts. 
% Different forgery techniques could produce distinct forgery artifacts, and detectors may become excessively specialized in detecting specific forgery methods, hindering their ability to generalize to unseen forgeries.
% In this paper, the key motivation is that representations encompassing a wider range of forgery types should enable the learning of a more adaptable decision boundary, thereby reducing overfitting to method-specific features. 
\begin{figure*}[t]
    \vspace{-1.0em}
    \centering
    \includegraphics[width=0.99\linewidth]{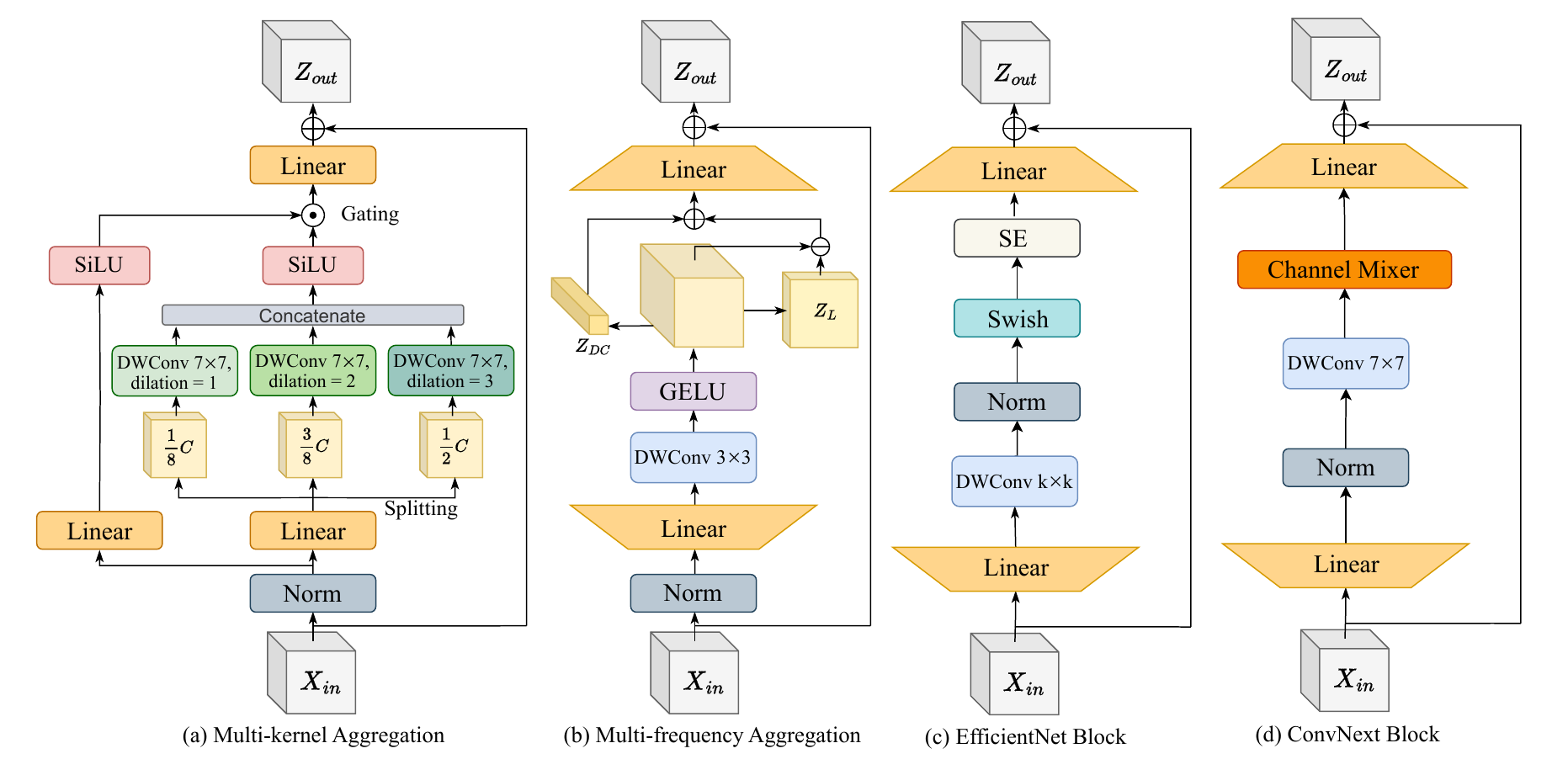}
    \vspace{-1.0em}
    \caption{(a) Structure of multi-kernel aggregation block as token mixer. (b) Structure of multi-frequency aggregation block as the channel mixer. (c) The basic building block of the EfficientNet model. (d) Structure of ConvNext block.}
    \label{fig:network}
\end{figure*}

% The MKA enhances local feature extraction by capturing features at different scales using depthwise separable convolutions with various dilation rates. The MFA improves the model's sensitivity to subtle forgery features by integrating features from different frequencies.
% Based on the above analysis and addressing the issues in the commonly used architectures for fake face detection, we have constructed a unique architecture specifically for fake face detection. 
% The basic modules of this architecture are shown in Figures \ref{fig}(c) and (d). We designed this architecture using modern deep neural network (DNN) principles and tailored specific basic modules for the task of fake face detection. These modules combine local relation learning, multi-scale feature fusion, and multi-frequency adaptive fusion.

% The basic block of this architecture is shown in Figure~\ref{fig:network}(c) (d) and combines local relation learning with multi-scale feature fusion. The multi-kernel aggregation module(MKA) captures features at different scales through depthwise separable convolutions with various dilation rates, enhancing the extraction of local features. The multi-frequency aggregation module(MFA) fuses features from different frequencies, increasing the model's sensitivity to subtle forgery features.

\subsection{Overview of MkfaNet}
Built upon modern ConvNets, we design a four-stage MkfaNet architecture as illustrated in 
Figure~\ref{fig:architecture}.
For stage $i$, the input image or feature is first fed into an embedding stem to regulate the resolutions and embed into $C_{i}$ dimensions. Assuming the input image in $H\times W$ resolutions, features of the four stages are in $\frac{H}{4}\times\frac{W}{4}$, $\frac{H}{8}\times\frac{W}{8}$, $\frac{H}{16}\times\frac{W}{16}$, and $\frac{H}{32}\times\frac{W}{32}$ resolutions respectively.
Then, the embedded feature flows into $N_{i}$ Mkfa Blocks, consisting of spatial and channel aggregation blocks, for multi-kernel feature and high-low frequency aggregation.

% fig: framework
\subsection{Multi-kernel Aggregator}
\label{procedure}
Model generative models are able to create extremely realistic fake human faces that are visually almost indistinguishable from real ones by learning from a vast amount of real facial data and thus simulating features such as lighting, texture, and shape of faces. In this context, traditional single-scale feature extraction methods struggle to detect these fake faces. This is mainly because such methods typically focus on features at a fixed scale, such as coarse patterns of edges or textures, and overlook the subtle changes and complex interactions across multiple scales, which are precisely what generation techniques excel at simulating.

Therefore, to effectively distinguish these high-quality fake images from real faces, a method capable of analyzing and identifying details at multiple levels is required. In this way, we propose MKA modules as a solution, which adaptively selects organ features extracted by multiple convolutions for modeling subtle facial differences between real and fake faces.
To elucidate the implementation details of the Multi-kernel Aggregator (MKA) module, as illustrated in Figure~\ref{fig:network}(a), we will delve into its architectural design, focusing on how it adaptively aggregates multi-level features to enhance the detection of key facial regions.
We represent this process as follows:
\begin{equation}
    Z = X + \mathrm{MKA}\big(\mathrm{Norm}(X)\big),
\end{equation}
where $\mathrm{MKA}(\cdot)$ denotes a multi-kernel gated aggregation module comprising the gating $\mathcal{F}_{\phi}(\cdot)$ and multi-kernel feature branch $\mathcal{G}_{\psi}(\cdot)$.

\paragraph{Multi-kernel feature extraction.}
To enable the model to perceive the multi-level features of the face images, we employ three different DWConv layers with dilation ratios $d\in \{1,2,3\}$ in parallel to capture low, middle, and high-order features: given the input feature $X\in \mathbb{R}^{C\times HW}$, the input is factorized into ${X}_l \in \mathbb{R}^{C_l \times HW}$, ${X}_m \in \mathbb{R}^{C_m \times HW}$, and ${X}_h \in \mathbb{R}^{C_h \times HW}$ along the channel dimension, where $C_l + C_m + C_h =C$; afterward,${X}_l$, ${X}_m$ and ${X}_h$ are assigned to $\mathrm{DW}_{7\times 7, d=1}$, $\mathrm{DW}_{7\times 7, d=2}$ and $\mathrm{DW}_{7\times 7, d=3}$, respectively. Finally, the output of ${X}_l$, ${X}_m$, and ${X}_h$ are concatenated to form multi-kernel feature, so that 
$Y_{C} = \mathrm{Concat}(Y_{l}, Y_{m}, Y_{h})$. 

\paragraph{Gated Aggregation.}
To \textit{adaptively} aggregate the extracted feature from the multi-kernel feature branch, and we employ SiLU activation in the gating branch, as $ \mathrm{SiLU}(x) = x\cdot \mathrm{Sigmoid}(x)$, which has been well-acknowledged as an advanced version of Sigmoid activation. 
SiLU has both the gating effect of Sigmoid and stable training characteristics, leading the final aggregated features as 
\begin{align}
    \label{eq:moga}
    Z &= \underbrace{\mathrm{SiLU}\big( \mathrm{Conv}_{1\times 1}(X) \big)}_{\mathcal{F}_{\phi}} \odot \underbrace{\mathrm{SiLU}\big( \mathrm{Conv}_{1\times 1}(Y_{C}) \big)}_{\mathcal{G}_{\psi}}.
\end{align}
% With the proposed MOSA blocks, the model captures more middle-order feature, as validated in Figure~\ref{fig:spatial_interaction}. 
\subsection{Multi-Frequency Aggregator}
Fig~\ref{fig:motivation}(b) shows the frequency domain analysis of the data, revealing significant differences in the distribution of high-frequency information between fake and real faces. Fake faces often appear unnatural in details such as skin texture and edge sharpness, resulting in a noticeably different distribution of features in the high-frequency region. Fig~\ref{fig:motivation}(c) illustrates the relative logarithmic amplitude of the Fourier-transformed data, with the color gradient from purple to yellow representing the transition from shallow to deep layers of the model. This gradient reveals how layers of different depths handle frequency information, providing visual evidence of the differences in frequency responses between real and fake faces.

It is evident that the shallow layers (purple) tend to capture high-frequency details related to texture and edges, while the deeper layers (yellow) strongly respond to low-frequency features, which are typically associated with the overall structure and shape of the image. At these levels, real and fake faces exhibit different frequency characteristics. Specifically, in the high-frequency details, fake faces often fail to perfectly replicate the high-frequency features of real faces due to technical limitations, resulting in anomalies or inconsistencies in the high-frequency region. This underscores the importance of addressing both low-frequency and high-frequency features in facial recognition.

We propose an MFA module that processes and reorganizes the direct current (DC) and high-frequency (HC) components of images independently, allowing the model to perform more refined and in-depth analysis at different frequency levels. Specifically, the MFA enhances the analysis of high-frequency details to identify unnatural textures and edges produced by generative models while integrating low-frequency information to maintain an understanding of the overall structure of the image. This approach not only strengthens the model's ability to detect flaws unique to forgery techniques but also improves its capacity to capture authentic features. As a result, the accuracy and robustness of facial authenticity recognition are significantly enhanced. By comprehensively analyzing features at different frequencies, the MFA helps the model better distinguish and recognize complex real and fake faces, effectively addressing the challenges posed by high-quality forgery techniques. The structure of the MFA module, as shown in Figure~\ref{fig:network}(b), can be formalized as follows:
\begin{equation}
\begin{aligned}
    Y &= \mathrm{GELU}\Big(\mathrm{DW_{3\times 3}}\big(\mathrm{Conv_{1\times 1}}(\mathrm{Norm}(X))\big)\Big),\\
    Z &= \mathrm{Conv_{1 \times 1}}\big(\mathrm{MF}(Y)\big) + X.
    \label{eq:ffn}
\end{aligned}
\end{equation}
where $\mathrm{MF}(\cdot)$ is a scaling technique that operates on feature maps by distinctively mixing information from different frequency bands. In specific, the input signal is firstly decomposed into its DC component and high-frequency components. Then, two sets of parameters are introduced to re-weight these components for each channel. The two-step processing reads as,
\begin{equation}
\begin{aligned}
    Y_{DC} &= z_{DC}\odot{Y},\\
    Y_{HC} &= Y - z_{L}\odot{Y},\\
    \mathrm{MF}(Y) &= Y_{DC} + \gamma\odot{Y_{HC}},
    \label{eq:mf}
\end{aligned}
\end{equation}
where $\gamma$ is the channel-wise scaling factor initialized as zeros. The calculation of DC and HC components is also computationally efficient, as it does not require explicit Fourier transforms. The DC component is calculated by averaging each feature map, while the HC component is obtained by subtracting the DC component from the original features. Specifically, 
$z_{DC}$ represents the spatial average, and $z_{L}$ represents the channel average.

\subsection{Discussion}
\subsubsection{Advantages over Classical CNN}
Currently, the commonly used architectures for fake face detection are XceptionNet~\cite{chollet2017xception} and EfficientNet~\cite{tan2019efficientnet}, which primarily focus on learning global features~\cite{wang2023dynamic,zhao2021multi}. The XceptionNet architecture uses depthwise separable convolution layers to build its structure. This design optimizes the learning of global features, making the model excel in recognizing overall structures and patterns in images. However, in tasks like fake face detection, which require fine analysis of local details, this tendency may limit the model's sensitivity to subtle facial expression differences and skin texture patterns. 

EfficientNet is an optimized convolutional neural network architecture that enhances model efficiency and feature representation by balancing adjustments to network depth, width, and resolution and integrating depthwise separable convolutions and Squeeze-and-Excitation (SE) blocks. Its architecture is illustrated in Figure~\ref{fig:network}(a). Although SE blocks improve the model's attention to features, this re-calibration is based on global information and cannot capture the specific local detail anomalies characteristic of forged faces. The performance gain of EfficientNet is attributing to its adjusting resolution. However, in fake face detection, sensitivity to subtle local feature variations is crucial. Simply relying on resolution adjustments is inadequate to counter high-quality face forgery techniques~\cite{deng2023cascaded}. 

Although Xception and EfficientNet employ innovative optimizations, adhering to traditional CNN design principles introduces certain disadvantages. Hierarchical feature extraction may not fully capture the global context and long-range dependencies in images, limiting their ability to handle complex data. Additionally, increasing the number of layers and network width enhances performance but significantly raises the model's parameter count and computational cost.
\subsubsection{Advantages over Modern DNN}
The design of modern deep neural networks (DNNs) offers significant advantages, including higher representational capacity and computational efficiency. By employing block-based designs combined with hierarchical and isotropic stages, they can effectively handle large-scale and complex datasets, capture long-range dependencies, and perform multi-scale feature extraction. Additionally, these networks can adaptively adjust the functionality and dimensions of each layer, providing greater flexibility to meet different task requirements, thereby significantly improving model performance while maintaining parameter efficiency.

ConvNext~\cite{cvpr2022convnext} is a modern convolutional neural network architecture. Figure~\ref{fig:network}(b) shows the architecture of its basic module. By using a channel mixer, ConvNext enhances the use of inter-channel information by mixing features from different channels, thus enriching feature representation. While the channel mixer improves the interaction between different channels, it may still lack sensitivity to local detail features, especially the subtle anomalies in forged faces. Therefore, although this model performs well in general image tasks, it may require further adjustments or integration with other mechanisms for specialized fake face detection tasks to better capture and analyze the inherent local and high-frequency detail features of forgery techniques.

Based on the above analysis and addressing the issues in commonly used architectures for fake face detection, we have constructed a unique architecture specifically for this task. This architecture is designed with modern deep neural network (DNN) principles in mind and incorporates multi-kernel feature adaptive fusion modules (MKA) and multi-frequency adaptive fusion modules (MFA), as shown in Fig~\ref{fig:network}(c) and (d), respectively.
The MKA module targets spatial context by adaptively selecting organ-specific features extracted through multiple convolutions to simulate the subtle facial differences between real and fake faces. The MFA module focuses on frequency components, processing different frequency bands by adaptively rebalancing high-frequency and low-frequency features.

\begin{table*}[!t]
    \centering
    % \vspace{-0.5em}
    \caption{Within-domain and cross-domain evaluations of various deepfake detectors and backbones using the AUC metric. All detectors are trained on FF-c23 and evaluated on other datasets. \textbf{Avg.} donates the average AUC for within-domain and cross-domain evaluations, and the best result for each group is highlighted in \textbf{bord}. $\dag$ represents our reproduced results, while DeepfakeBench provides others.}
    \vspace{-0.5em}
    \setlength{\tabcolsep}{0.7mm}
\resizebox{1.0\linewidth}{!}{
    \begin{tabular}{lccc|ccccccc|ccccccc}
\toprule
\bf{Type} & \bf{Detector}                              & \bf{Backbone}      & \bf{\# Param.} & \multicolumn{7}{c|}{\bf{Within Domain Evaluation}}                                              & \multicolumn{7}{c}{\bf{Cross Domain Evaluation}}                                                \\
          &                                            &                    & (M)            & FF-c23      & FF-c40      & FF-DF       & FF-F2F      & FF-FS       & FF-NT       & Avg.        & CDFv1       & CDFv2       & DF-1.0      & DFD         & DFDC        & DFDCP       & Avg.        \\ \hline
Naive     & Meso4~\cite{afchar2018mesonet}             & MesoNet            & 0.03           & 0.6077      & 0.5920      & 0.6771      & 0.6170      & 0.5946      & 0.5701      & 0.6097      & 0.7358      & 0.6091      & 0.9113      & 0.5481      & 0.5560      & 0.5994      & 0.6599      \\
Naive     & MesoIncep~\cite{afchar2018mesonet}         & MesoNet            & 0.03           & 0.7583      & 0.7278      & 0.8542      & 0.8087      & 0.7421      & 0.6517      & 0.7571      & 0.7366      & 0.6966      & 0.9233      & 0.6069      & 0.6226      & \bf{0.7561} & 0.7237      \\
Spatial   & Capsule~\cite{nguyen2019capsule}           & Capsule            & 4.0            & 0.8421      & 0.7040      & 0.8669      & 0.8634      & 0.8734      & \bf{0.7804} & 0.8217      & \bf{0.7909} & 0.7472      & 0.9107      & 0.6841      & 0.6465      & 0.6568      & 0.7394      \\
Naive     & CNN-Aug~\cite{wang2020cnn}                 & ResNet-34          & 22             & 0.8493      & 0.7846      & \bf{0.9048} & 0.8788      & 0.9026      & 0.7313      & 0.8419      & 0.7420      & 0.7027      & 0.7993      & 0.6464      & 0.6361      & 0.6170      & 0.6906      \\
\rowcolor{gray94} Naive  & \textbf{MkfaNet}                       & \textbf{MkfaNet-T} & 5.2            & \bf{0.8506} & \bf{0.7879} & 0.8982      & \bf{0.8823} & \bf{0.9037} & 0.7796      & \bf{0.8504} & 0.7881      & \bf{0.7486} & \bf{0.9245} & \bf{0.6883} & \bf{0.6477} & 0.7428      & \bf{0.7567} \\ \hline
Naive     & CNN-Aug~\cite{wang2020cnn}                 & ResNet-50$^\dag$   & 26             & 0.8925      & 0.7956      & 0.9258      & 0.9135      & 0.9252      & 0.7828      & 0.8726      & 0.7608      & 0.7591      & 0.8143      & 0.6995      & 0.6661      & 0.6245      & 0.7207      \\
Naive     & Xception~\cite{rossler2019faceforensics++} & Xception           & 23             & 0.9637      & 0.8261      & 0.9799      & 0.9785      & 0.9833      & 0.9385      & 0.9450      & 0.7794      & 0.7365      & 0.8341      & 0.8163      & 0.7077      & 0.7374      & 0.7686      \\
Naive     & Efficient~\cite{tan2019efficientnet}       & Efficient-B4       & 19             & 0.9567      & 0.8150      & 0.9757      & 0.9758      & 0.9797      & 0.9308      & 0.9389      & 0.7909      & 0.7487      & 0.8330      & 0.8148      & 0.6955      & 0.7283      & 0.7685      \\
\rowcolor{gray94} Naive     & Swin~\cite{iccv2021swin}                   & Swin-T$^\dag$      & 28             & 0.9630      & 0.8278      & 0.9802      & 0.9783      & 0.9826      & 0.9375      & 0.9449      & 0.7863      & 0.7476      & 0.8384      & 0.8028      & 0.7053      & 0.7339      & 0.7691      \\
\rowcolor{gray94} Naive     & ConvNeXt~\cite{cvpr2022convnext}           & ConvNeXt-T$^\dag$  & 29             & 0.9644      & 0.8287      & 0.9796      & 0.9801      & 0.9840      & 0.9393      & 0.9460      & 0.7837      & 0.7491      & 0.8425      & 0.8102      & 0.7075      & 0.7366      & 0.7716      \\
\rowcolor{gray94} Naive     & \textbf{MkfaNet}                           & \textbf{MkfaNet-S} & 20             & \bf{0.9671} & \bf{0.8315} & \bf{0.9826} & \bf{0.9820} & \bf{0.9849} & \bf{0.9428} & \bf{0.9485} & \bf{0.7946} & \bf{0.7538} & \bf{0.8785} & \bf{0.8166} & \bf{0.7127} & \bf{0.7413} & \bf{0.7829} \\ \hline
Frequency & F3Net~\cite{qian2020thinking}              & Xception           & 23             & 0.9635      & 0.8271      & 0.9793      & 0.9796      & 0.9844      & 0.9354      & 0.9449      & 0.7769      & 0.7352      & 0.8431      & 0.7975      & 0.7021      & 0.7354      & 0.7650      \\
Frequency & SPSL~\cite{liu2021spatial}                 & Xception           & 23             & 0.9610      & 0.8174      & 0.9781      & 0.9754      & 0.9829      & 0.9299      & 0.9408      & \bf{0.8150} & 0.7650      & 0.8767      & 0.8122      & 0.7040      & 0.7408      & 0.7856      \\
Frequency & SRM~\cite{luo2021generalizing}             & Xception           & 23             & 0.9576      & 0.8114      & 0.9733      & 0.9696      & 0.9740      & 0.9295      & 0.9359      & 0.7926      & 0.7552      & 0.8638      & 0.8120      & 0.6995      & 0.7408      & 0.7773      \\
Spatial   & FWA~\cite{li2018exposing}                  & Xception           & 23             & 0.8765      & 0.7357      & 0.9210      & 0.9000      & 0.8843      & 0.8120      & 0.8549      & 0.7897      & 0.6680      & \bf{0.9334} & 0.7403      & 0.6132      & 0.6375      & 0.7303      \\
Spatial   & X-ray~\cite{li2020face}                    & HRNet              & 22             & 0.9592      & 0.7925      & 0.9794      & \bf{0.9872} & 0.9871      & 0.9290      & 0.9391      & 0.7093      & 0.6786      & 0.5531      & 0.7655      & 0.6326      & 0.6942      & 0.6722      \\
Spatial   & FFD~\cite{dang2020detection}               & Xception           & 22             & 0.9624      & 0.8237      & 0.9803      & 0.9784      & 0.9853      & 0.9306      & 0.9434      & 0.7840      & 0.7435      & 0.8609      & 0.8024      & 0.7029      & 0.7426      & 0.7727      \\
Spatial   & CORE~\cite{ni2022core}                     & Xception           & 22             & 0.9638      & 0.8194      & 0.9787      & 0.9803      & 0.9823      & 0.9339      & 0.9431      & 0.7798      & 0.7428      & 0.8475      & 0.8018      & 0.7049      & 0.7341      & 0.7685      \\
Spatial   & UCF~\cite{yan2023ucf}                      & Xception           & 47             & 0.9705      & 0.8399      & 0.9883      & 0.9840      & 0.9896      & 0.9441      & 0.9527      & 0.7793      & 0.7527      & 0.8241      & 0.8074      & 0.7191      & 0.7594      & 0.7737      \\
\rowcolor{gray94} Spatial   & FFD~\cite{dang2020detection} & \textbf{MkfaNet-S} & 20             & \bf{0.9829} & \bf{0.8475} & \bf{0.9916} & 0.9869      & \bf{0.9937} & \bf{0.9524} & \bf{0.9591} & 0.8065      & \bf{0.7679} & 0.8930      & \bf{0.8194} & \bf{0.7258} & \bf{0.7652} & \bf{0.7963} \\
\bottomrule
    \end{tabular}
    }
    \label{tab:deepfakebench}
    % \vspace{-1.0em}
\end{table*}

\section{Experiments}
\label{sec:exp}
\subsection{Settings}
\paragraph{Datasets.}
To evaluate the performances and generalization abilities of our proposed backbone, we follow DeepfakeBench \cite{yan2023deepfakebench} to conduct comparison and analysis experiments on seven commonly used deepfake detection datasets:
FaceForensics++ (FF++)~\cite{rossler2019faceforensics++}, CelebDF-v1 (CDFv1)~\cite{li2019celeb}, CelebDF-v2 (CDFv2)~\cite{li2019celeb}, DeepFakeDetection (DFD)~\cite{2021dfd}, DeepFake Detection Challenge Preview (DFDC-P)~\cite{dolhansky2019deepfake}, DeepFake Detection Challenge (DFDC)~\cite{dolhansky2020deepfake}, and DeeperForensics-1.0 (DF-1.0)~\cite{jiang2020deeperforensics}. 
Specifically, FF++ is a large-scale database with 1.8 million forged images that contains 4 types of manipulation methods, including Deepfakes (FF-DF)~\cite{2021deepfake}, Face2Face (FF-F2F)~\cite{thies2016face2face}, FaceSwap (FF-FS)~\cite{2021faceswap}, and NeuralTextures (FF-NT)~\cite{thies2019deferred}. Note that we use the lightly compressed (c23) version of FF++ as the default training data, whereas two other compressed versions of FF++ are raw and heavily compressed (c40), while others are used as testing datasets. We also adopt the full data pre-processing workflow proposed in DeepfakeBench and use the fixed training and testing resolutions of $256\times 256$ for the cropped face images.

\paragraph{Implementation Details.}
For a fair comparison, we consider three types of detectors in DeepfakeBench, as detailed in Table~\ref{tab:deepfakebench}: \textbf{(1) Naive detectors} that combine a backbone and binary classifier without introducing manually designed features. Both classical CNNs (\textit{e.g.,} ResNet~\cite{he2016deep} and EfficientNet~\cite{tan2019efficientnet}) and modern architectures (\textit{e.g.,} Swin Transformer~\cite{iccv2021swin} and ConvNeXt~\cite{cvpr2022convnext}) are compared. \textbf{(2) Spatial detectors} that build upon the backbone and further utilize spatial features with manual-designed algorithms. \textbf{(3) Frequency detectors} focus on exploring frequency components and artifacts to detect forgeries.
As for training settings, we train detectors with classical CNNs as the backbone by Adam optimizer~\cite{iclr2015adam} with a learning rate of $2\times 10^{-4}$ and a batch size of 32 for all experiments. Following ConvNeXt~\cite{cvpr2022convnext}, we employ AdamW optimizer~\cite{iclr2019AdamW} to fine-tune detectors with modern backbones using the learning rate of $5\times 10^{-4}$ and the batch size of 256. Pre-trained weights of the backbone on ImageNet-1K~\cite{cvpr2009imagenet} will be used if feasible, and our MkfaNet adopts the same pre-training setting as ConvNeXt-T~\cite{cvpr2022convnext} on ImageNet-1K. We also apply data augmentations, including image compression, horizontal flip, rotation, Gaussian blur, and random brightness contrast.
% \paragraph{Evaluation Metrics.} 
As for evaluation metrics, we conduct the \textbf{frame-level Area Under Curve (AUC)} to compare our proposed MkfaNet with existing works, where the mean result over three trials is reported.

\subsection{Comparison Results}
\label{sec:comparison}
As shown in Table~\ref{tab:deepfakebench}, we conduct within-domain and cross-domain evaluations for three versions of our MkfaNet, \textit{i.e.,} the lightweight detectors, the naive detectors compared to various backbones, and the advanced detectors with prior knowledge from spatial or frequency domains. 
cross-domain evaluation involves testing the model on different datasets.

\begin{figure*}[!t]
\centering
    \includegraphics[width=0.95\linewidth]{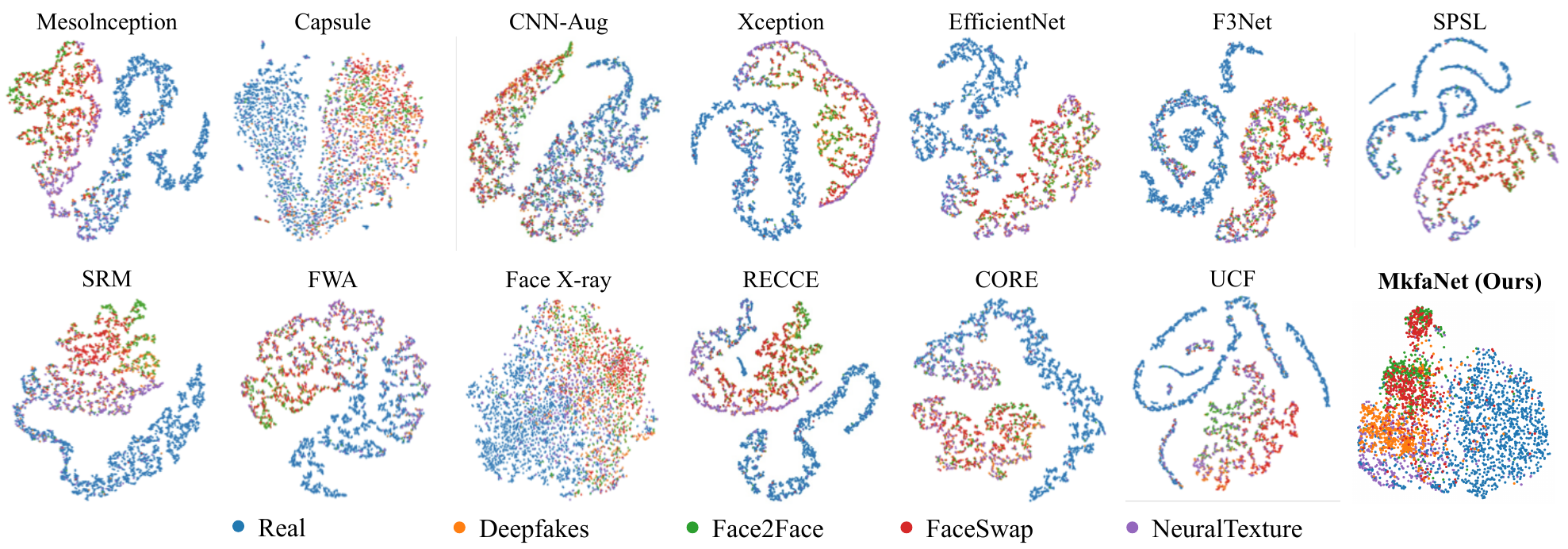} 
    \vspace{-0.5em}
    \caption{Visualization of latent embedding of detectors with t-SNE~\cite{jmlr2008tsne} on FF++ (c23) according to DeepfakeBench~\cite{yan2023deepfakebench}.}
    \label{fig:tsne}
    \vspace{-1.0em}
\end{figure*}

\vspace{-0.5em}
\paragraph{Within-domain Evaluations.}
We first conduct within-domain evaluations to verify the performances of detectors within the same dataset following DeepfakeBench~\cite{yan2023deepfakebench}. Table~\ref{tab:deepfakebench} (middle columns) shows MkfaNet variants achieve the best average results on six within-domain datasets. As for the lightweight detectors, MkfaNet-T significantly outperforms MesoNet~\cite{afchar2018mesonet} and CapsuleNet~\cite{nguyen2019capsule} with similar parameters by 9.33\% and 2.87\% AUC while outperforming CNN-Aug~\cite{wang2020cnn} using only a quarter of the parameters of ResNet-34~\cite{he2016deep}.
When compared to naive detectors with around 20M parameters, the modern networks (Swin-T~\cite{iccv2021swin} and ConvNeXt-T~\cite{cvpr2022convnext}) consistently improve the classical CNNs (ResNet-50~\cite{he2016deep}, Xception~\cite{chollet2017xception}, and EfficientNet-B4~\cite{tan2019efficientnet}), \textit{e.g.,} ConvNeXt-T outperforms ResNet-50 and EfficientNet-B4 by 10.29\% and 0.77\% AUC on FF-c23, which might attribute to the Metaformer macro design~\cite{tpami2024MetaFormer} and more parameters. Meanwhile, our proposed MkfaNet-S significantly improves both classical CNNs and modern networks with efficient usage of parameters, \textit{e.g.,} MkfaNet-S yields 94.85\% AUC and around 0.25$\sim$8.0\% performance gains in average comparing to previous backbones.
When employing larger backbone encoders with manually-designed features, FFD~\cite{dang2020detection} with MkfaNet-S significantly outperforms frequency detectors (F3Net~\cite{qian2020thinking}, SPSL~\cite{liu2021spatial}, and SRM~\cite{luo2021generalizing}) and spatial detectors (FWA~\cite{li2018exposing}, X-ray~\cite{li2020face}, and CORE~\cite{ni2022core}) with Xception and HRNet~\cite{cvpr2019hrnet} backbones in the similar parameter scale, while even yielding better results than UCF~\cite{yan2023ucf} with a larger Xception backbone.

\vspace{-0.5em}
\paragraph{Cross-domain Evaluations.}
Then, we evaluate detectors on different datasets without further fine-tuning, which reflects the generalization and robustness of the compared detectors. As shown in Table~\ref{tab:deepfakebench} (right columns), all models suffer performance decreases because of the challenging domain gap. Surprisingly, our proposed MkfaNet variants achieve the best average results and form greater performance gains over existing methods, \textit{e.g.,} Naive detector with MkfaNet-S outperforms Xception and ConvNeXt-T by 1.43\% and 1.13\% average AUC, indicating that MkfaNet might learn more common and robust features. We verify this hypothesis in Sec.~\ref{sec:ablation} with visualizations.

\begin{figure}[!t]
    \vspace{-0.5em}
    \centering
    \includegraphics[width=1.0\linewidth]{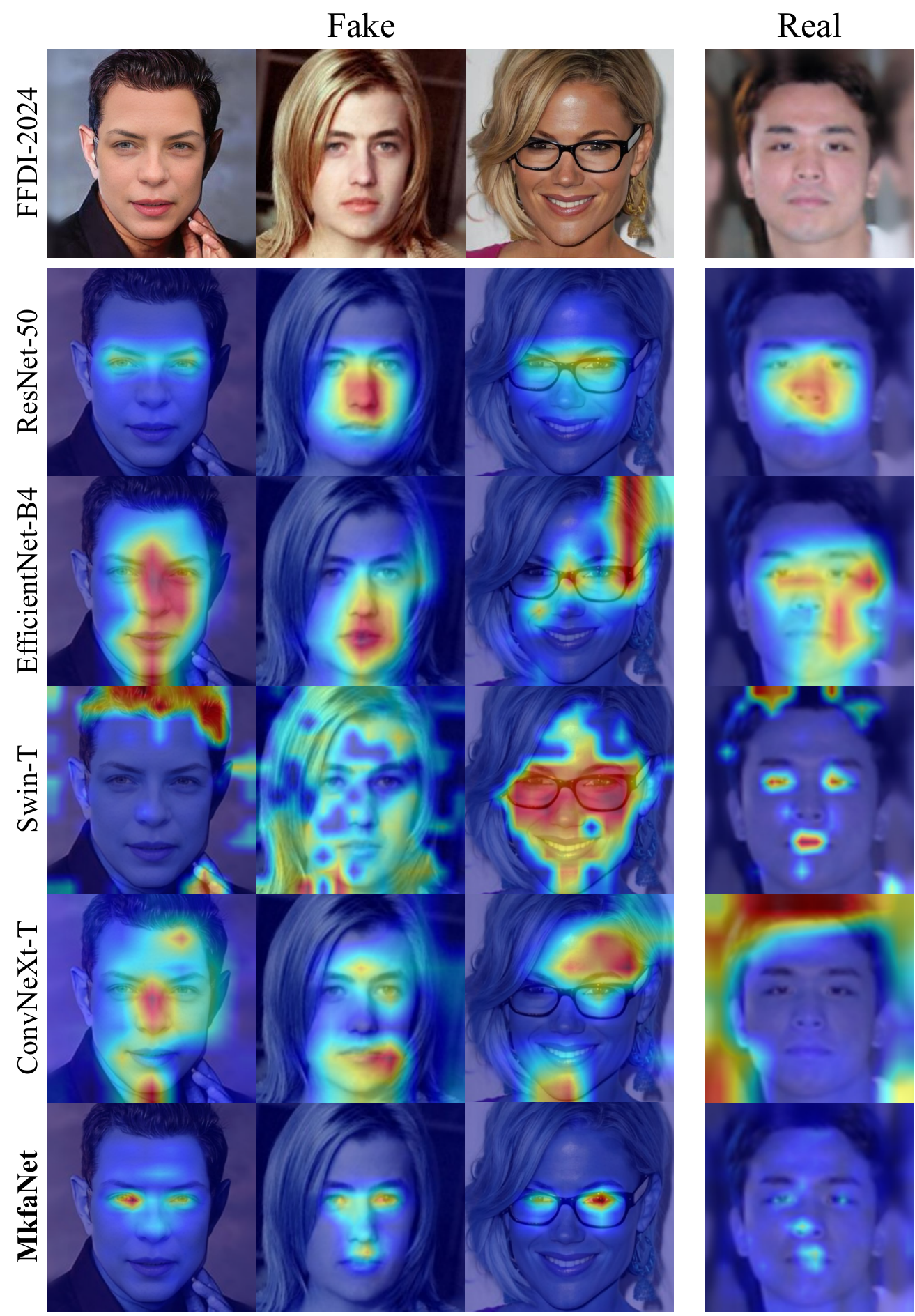}
    \vspace{-1.25em}
    \caption{
    Grad-CAM activation maps \cite{cvpr2017grad} of fake and real images in the validation set of FFDI-2024 as cross-domain evaluation. Compare the naive detector with different backbones with ours. As for fake images, classical CNNs like ResNet-50 show robust but coarse localization of human faces, while modern architectures like Swin-T can activate some semantic features. Out MkfaNet not only exhibits precise localization of discriminative organs but also tells the difference between fake and real faces.
    }
    \label{fig:gradcam} 
    \vspace{-0.5em}
\end{figure}

\begin{table}[ht]
    \setlength{\tabcolsep}{1.5mm}
    \centering
\resizebox{0.99\linewidth}{!}{
    \begin{tabular}{cc|cc}
    \toprule
\bf{Block} & \bf{Module}                                & FF-c23      & \# Param. \\
           &                                            & (AUC)       & (M)       \\ \hline
ResNet     & Bottleneck                                 & 0.8437      & 25.6      \\
ConvNeXt   & DWConv$7\times 7$+FFN                      & 0.8856      & 28.5      \\ \hline
           & \cellcolor{gray94}Gating Branch            & 0.8819      & 18.5      \\
MKA        & +DWConv7$\times 7$                         & 0.8932      & 18.7      \\
           & \cellcolor{gray94}+Multi-DWConv$7\times 7$ & \bf{0.9015} & 19.0      \\ \hline
           & \cellcolor{gray94}DWConv$3\times 3$+FFN    & 0.9093      & 19.2      \\
MFA        & +SE                                        & 0.9150      & 21.5      \\
           & \cellcolor{gray94}+MF                      & \bf{0.9176} & 19.8      \\
    \bottomrule
    \end{tabular}
    }
    \vspace{-0.5em}
    \caption{Ablation of designed modules on FF-c23. The module without ``+" denotes the baseline modules, while those with ``+" are added to the baseline (using \textcolor{gray}{gray} backgrounds).
    }
    \label{tab:ablation}
    % \vspace{-1.0em}
\end{table}

\subsection{Ablation and Analysis}
\label{sec:ablation}
\paragraph{Ablation studies of network modules.}
We first ablate the designed modules in MkfaNet with a simplified experimental setting, \textit{i.e.,} training and evaluation on FF-c23 without using ImageNet-1K pre-trained weights. We take ConvNeXt-T~\cite{cvpr2022convnext} as the baseline for MkfaNet, which outperforms the classical bottleneck in ResNet-50~\cite{he2016deep} in Table~\ref{tab:ablation}. As for the proposed \textbf{Multi-kernel Aggregator} (MKA) block, using the \textbf{Gating Branch} in Eq.~\ref{eq:moga} can yield similar performances as ConvNeXt-T with around 10M fewer parameters and using \textbf{Multi-DWConv$7\times 7$} with dilated ratios in $(1, 2, 3)$ aggregates contextualized patterns and improves the performances. As for the \textbf{Multi-Frequency Aggregator} (MFA) block, adding a Squeeze-and-excitation (SE) module~\cite{hu2018senet} to \textbf{DWConv$3\times 3$ + FFN} is equivalent to the EfficientNet block~\cite{tan2019efficientnet}, which requires numerous parameters for performance gains. Our proposed \textbf{MF} module in Eq.~\ref{eq:mf} brings better AUC than the SE module while using fewer parameters.

\paragraph{Visualization analysis.}
We then evaluate the learned features of MkfaNet-S by two visualizations. As shown in Figure~\ref{fig:tsne}, the representations of various detectors are visualized by t-SNE~\cite{jmlr2008tsne} on FF++ (c23) dataset with 5000 randomly selected samples following DeepfakeBench, where four forgery types (Deepfakes, Face2Face, FaceSwap, and NeuralTextures) in FF++ are considered. The representations of real and fake samples are more separable in MkfaNet-S than in previous works, while four different forgeries are also discriminative by MkfaNer-S.
It indicates that MkfaNet can capture common features rather than over-fitting the training dataset.
Meanwhile, we further investigate the spatial features learned by MkfaNet with Grad-CAM~\cite{cvpr2017grad} visualization in Figure~\ref{fig:gradcam}. We consider cross-domain evaluation samples from FFDI-2024\footnote{The latest Global Multimedia Deepfake Detection competition at \href{https://www.kaggle.com/competitions/multi-ffdi/data}{https://www.kaggle.com/competitions/multi-ffdi/data}.}, and compare with various backbone architectures. Figure~\ref{fig:gradcam} shows that MkfaNet-S precisely and consistently locates organs to determine fake or real faces, while other backbones sometimes extract irrelevant regions, which might deteriorate the generalization and robustness of forgery detection.

\section{Conclusion}
In this paper, we introduce MkfaNet, a novel backbone network specifically designed for face forgery detection. It combines two core modules: the Multi-Kernel Aggregator (MKA) and the Multi-Frequency Aggregator (MFA), which effectively enhance the ability to distinguish between real and forged facial features. The MKA module targets spatial context by adaptively selecting organ-specific features extracted through multiple convolutions to simulate the subtle facial differences between real and fake faces. The MFA module focuses on frequency components, processing different frequency bands by adaptively rebalancing high-frequency and low-frequency features. This innovative approach not only significantly improves the accuracy of forgery detection but also enhances the model's capability to handle complex facial data, making it a powerful tool for combating advanced forgery techniques in the future.

%%%%%%%%% REFERENCES
{\small
\bibliographystyle{ieee_fullname}
\bibliography{main}
}

%%%%%%%%% Appendix
\renewcommand\thefigure{A\arabic{figure}}
\renewcommand\thetable{A\arabic{table}}
\setcounter{table}{0}
\setcounter{figure}{0}

\newpage
\appendix

\section*{\large{Appendix}}
The appendix section provides details of the MkfaNet network and experiments.

\section{Implementation Details}
\subsection{Network Configurations}
We provide detailed architecture configurations of MkfaNet variants in Table~\ref{tab:app_network}, where we scale the embedding dimensions and the number of blocks for each stage:
(1) MkfaNet-Tiny with embedding dimensions of $\{32, 64, 128, 256\}$ is designed for lightweight deepfake detection scenarios, exhibiting competitive parameter numbers around 5M parameters;
(2) MkfaNet-Small utilizes embedding dimensions of $\{64, 128, 320, 512\}$ in comparison to other prevailing modern architectures~\cite{cvpr2022convnext} around 25M parameters.

\begin{table}[ht]
    \centering
    % \vspace{-0.5em}
    \caption{Architecture configurations of MkfaNet variants.}
    \vspace{-0.5em}
    \setlength{\tabcolsep}{1.1mm}
\resizebox{0.92\linewidth}{!}{
    \begin{tabular}{ccc|cc}
\toprule
\multicolumn{1}{c|}{Stage} & \multicolumn{1}{c|}{Output} & Layer & \multicolumn{2}{c}{MkfaNet} \\ \cline{4-5} 
\multicolumn{1}{c|}{} & \multicolumn{1}{c|}{Size} & Settings & \multicolumn{1}{c|}{~~~Tiny~~~} & Small \\ \hline
\multicolumn{1}{c|}{\multirow{4}{*}{S1}} & \multicolumn{1}{c|}{\multirow{4}{*}{$\frac{H\times W}{4\times 4}$}} & Stem & \multicolumn{2}{c}{
\small{\begin{tabular}[c]{@{}c@{}}$\rm{Conv}_{3\times 3},~\rm{stride}~2, C/2$ \\ $\rm{Conv}_{3\times 3},~\rm{stride}~2, C$\end{tabular}}
} \\ \cline{3-5} 
\multicolumn{1}{c|}{} & \multicolumn{1}{c|}{} & Embed. Dim. & \multicolumn{1}{c|}{32} & 64 \\ \cline{3-5} 
\multicolumn{1}{c|}{} & \multicolumn{1}{c|}{} & \# Moga Block & \multicolumn{1}{c|}{3} & 2 \\ \cline{3-5} 
\multicolumn{1}{c|}{} & \multicolumn{1}{c|}{} & MLP Ratio & \multicolumn{2}{c}{8} \\ \hline
\multicolumn{1}{c|}{\multirow{4}{*}{S2}} & \multicolumn{1}{c|}{\multirow{4}{*}{$\frac{H\times W}{8\times 8}$}} & Stem & \multicolumn{2}{c}{$\rm{Conv}_{3\times 3}, \rm{stride}~2$} \\ \cline{3-5} 
\multicolumn{1}{c|}{} & \multicolumn{1}{c|}{} & Embed. Dim. & \multicolumn{1}{c|}{64} & 128 \\ \cline{3-5} 
\multicolumn{1}{c|}{} & \multicolumn{1}{c|}{} & \# Moga Block & \multicolumn{1}{c|}{3} & 3 \\ \cline{3-5} 
\multicolumn{1}{c|}{} & \multicolumn{1}{c|}{} & MLP Ratio & \multicolumn{2}{c}{8} \\ \hline
\multicolumn{1}{c|}{\multirow{4}{*}{S3}} & \multicolumn{1}{c|}{\multirow{4}{*}{$\frac{H\times W}{16\times 16}$}} & Stem & \multicolumn{2}{c}{$\rm{Conv}_{3\times 3},~\rm{stride}~2$} \\ \cline{3-5} 
\multicolumn{1}{c|}{} & \multicolumn{1}{c|}{} & Embed. Dim. & \multicolumn{1}{c|}{128} & 320 \\ \cline{3-5} 
\multicolumn{1}{c|}{} & \multicolumn{1}{c|}{} & \# Moga Block & \multicolumn{1}{c|}{12} & 10 \\ \cline{3-5} 
\multicolumn{1}{c|}{} & \multicolumn{1}{c|}{} & MLP Ratio & \multicolumn{2}{c}{4} \\ \hline
\multicolumn{1}{c|}{\multirow{4}{*}{S4}} & \multicolumn{1}{c|}{\multirow{4}{*}{$\frac{H\times W}{32\times 32}$}} & Stem & \multicolumn{2}{c}{$\rm{Conv}_{3\times 3},~\rm{stride}~2$} \\ \cline{3-5} 
\multicolumn{1}{c|}{} & \multicolumn{1}{c|}{} & Embed. Dim. & \multicolumn{1}{c|}{256} & 512 \\ \cline{3-5} 
\multicolumn{1}{c|}{} & \multicolumn{1}{c|}{} & \# Moga Block & \multicolumn{1}{c|}{2} & 2 \\ \cline{3-5} 
\multicolumn{1}{c|}{} & \multicolumn{1}{c|}{} & MLP Ratio & \multicolumn{2}{c}{4} \\ \hline
\multicolumn{3}{c|}{Parameters (M)} & \multicolumn{1}{c|}{5.2} & 19.8 \\
\bottomrule
    \end{tabular}
    }
    \label{tab:app_network}
    % \vspace{-1.0em}
\end{table}

\subsection{Datasets}
As shown in Table~\ref{tab:app_dataset}, we provide detailed information for the used deepfake detection datasets. Among them, FaceForensics++ (FF++)~\cite{rossler2019faceforensics++} can be divided into five subsets for training and within-domain evaluations, including FF-DF, FF-F2F, FF-FS, FF-NT, and FF-all. Each subset corresponds to a combination of deepfake and real videos from YouTube. Other datasets are used for cross-domain evaluation.
We follow pre-processing scripts in DeepfakeBench~\cite{yan2023deepfakebench} to prepare the training and testing datasets, which incorporates four major steps, including face detection, face cropping, face alignment, and various other pre-processing operations. All images are aligned, cropped, and resized to $256\times 256$ resolutions.

\begin{table}[ht]
    \centering
    % \vspace{-0.5em}
    \caption{Summary of used deepfake detection datasets.}
    \vspace{-0.5em}
    \setlength{\tabcolsep}{0.6mm}
\resizebox{1.0\linewidth}{!}{
    \begin{tabular}{l|c|cccccc}
\toprule
Dataset                                & Domain & Real   & Fake    & Total   & Rights  & Total    & Synthesis \\
                                       &        & Videos & Videos  & Videos  & Cleared & Subjects & Methods   \\ \hline
FF++~\cite{rossler2019faceforensics++} & Within & 1000   & 4000    & 5000    & NO      & N/A      & 4         \\
FaceShifter~\cite{li2019faceshifter}   & Cross  & 1000   & 1000    & 2000    & NO      & N/A      & 1         \\
DFD~\cite{2021dfd}                     & Cross  & 363    & 3000    & 3363    & YES     & 28       & 5         \\
DFDC-P~\cite{dolhansky2019deepfake}    & Cross  & 1131   & 4119    & 5250    & YES     & 66       & 2         \\
DFDC~\cite{dolhansky2020deepfake}      & Cross  & 23,654 & 104,500 & 128,154 & YES     & 960      & 8         \\
CelebDF-v1~\cite{li2019celeb}          & Cross  & 408    & 795     & 1203    & NO      & N/A      & 1         \\
CelebDF-v2~\cite{li2019celeb}          & Cross  & 590    & 5639    & 6229    & NO      & 59       & 1         \\
DF-1.0~\cite{jiang2020deeperforensics} & Cross  & 50,000 & 10,000  & 60,000  & YES     & 100      & 1         \\
\bottomrule
    \end{tabular}
    }
    \label{tab:app_dataset}
    % \vspace{-1.0em}
\end{table}

\subsection{Experimental Settings}
As for training settings, we utilize Adam optimizer~\cite{iclr2015adam} with a basic learning rate of $2\times 10^4$, a batch size of 32 by default. 
For the naive detectors with modern backbones (\textit{e.g.}, ConvNeXt-T~\cite{cvpr2022convnext}), we utilize AdamW optimizer~\cite{iclr2019AdamW} to fine-tune detectors with the learning rate of $5\times 10^{-4}$ and the batch size of 256 according to their official settings on ImageNet-1K~\cite{cvpr2009imagenet}. As initializing the parameters with pre-training provides better performance, we also conduct 300-epoch pre-training on ImageNet-1K for MkfaNet-T/S, as shown in Table~\ref{tab:app_pretraining}.
As for evaluation protocols, we consider the average Area Under the Curve (AUC) over three runs as our primary metric. Furthermore, it is important to note that the validation set is not utilized in our experiments.

\begin{table}[ht]
    \centering
    % \vspace{-0.5em}
    \caption{Hyper-parameters and training recipes for ImageNet-1K of Swin-T, ConvNeXt-T, and our proposed MkfaNet-T/S.}
    \vspace{-0.5em}
    \setlength{\tabcolsep}{1.3mm}
\resizebox{0.95\linewidth}{!}{
    \begin{tabular}{l|c|c|c}
\toprule
Configuration              & Swin         & ConvNeXt     & MkfaNet      \\
                           & Tiny         & Tiny         & Tiny/Small   \\ \hline
Input resolution           & 224$^2$      & 224$^2$      & 224$^2$      \\
Epochs                     & 300          & 300          & 300          \\
Batch size                 & 1024         & 4096         & 4096         \\
Optimizer                  & AdamW        & AdamW        & AdamW        \\
AdamW $(\beta_1, \beta_2)$ & $0.9, 0.999$ & $0.9, 0.999$ & $0.9, 0.999$ \\
Learning rate              & 0.001        & 0.004        & 0.004        \\
Learning rate decay        & Cosine       & Cosine       & Cosine       \\
Weight decay               & 0.05         & 0.05         & 0.04/0.05    \\
Warmup epochs              & 20           & 20           & 20           \\
Label smoothing $\epsilon$ & 0.1          & 0.1          & 0.1          \\
Stochastic Depth           & \cmarkg      & \cmarkg      & \cmarkg      \\
Rand Augment               & 9/0.5        & 9/0.5        & 9/0.5        \\
Repeated Augment           & \cmarkg      & \xmarkg      & \xmarkg      \\
Mixup $\alpha$             & 0.8          & 0.8          & 0.1/0.8      \\
CutMix $\alpha$            & 1.0          & 1.0          & 1.0          \\
Erasing prob.              & 0.25         & 0.25         & \xmarkg/0.25 \\
ColorJitter                & \xmarkg      & \xmarkg      & \xmarkg      \\
Gradient Clipping          & \cmarkg      & \xmarkg      & \xmarkg      \\
EMA decay                  & \cmarkg      & \cmarkg      & \cmarkg      \\
Test crop ratio            & 0.875        & 0.875        & 0.90         \\
\bottomrule
    \end{tabular}
    }
    \label{tab:app_pretraining}
    % \vspace{-1.0em}
\end{table}

\end{document}